%% file: main.tex
\documentclass[letterpaper, 10 pt, conference]{ieeeconf}  

\IEEEoverridecommandlockouts                              

\usepackage{multicol}
\usepackage[bookmarks=true]{hyperref}

\usepackage{amsmath}
\usepackage{graphicx}
\usepackage[colorinlistoftodos]{todonotes}

\usepackage{amssymb}  

\usepackage{algorithm}
\usepackage{algorithmic}

\usepackage{caption}
\usepackage{multirow}
\usepackage{verbatim}
\newtheorem{theorem}{Theorem}[section]
\newtheorem{lemma}[theorem]{Lemma}

\usepackage[caption=false,subrefformat=parens,labelformat=parens]{subfig}
\usepackage{esint}
\usepackage{url}
\usepackage{color}

\title{\LARGE \bf
Efficient Probabilistic Collision Detection for \\Non-Gaussian Noise Distributions
}

\author{Jae Sung Park$^{1}$ and Dinesh Manocha$^{2}$%
\thanks{$^{1}$ Jae Sung Park with the Department of Computer Science, University of North Carolina at Chapel Hill, NC, USA {\tt\small jaesungp@cs.unc.edu}}%
\thanks{$^{2}$ Dinesh Manocha is with Department of Computer Science and Electrical \& Computer Engineering, University of Maryland at College Park, MD, USA {\tt\small dmanocha@umd.edu}}%
}

\begin{document}

\maketitle
\thispagestyle{empty}
\pagestyle{empty}

\begin{abstract}
We present an efficient algorithm to compute tight upper bounds of collision probability between two objects with positional uncertainties, whose error distributions are represented with  non-Gaussian forms. Our approach can handle noisy datasets from depth sensors, whose distributions may correspond  to  Truncated Gaussian, Weighted Samples, or Truncated Gaussian Mixture Model.  We derive tight probability bounds for convex shapes and extend them to   non-convex shapes using hierarchical representations. We highlight the benefits of our approach over prior  probabilistic collision detection algorithms in terms of tighter bounds ($10$x) and improved running time ($3$x). Moreover, we use our tight bounds to design an efficient and accurate motion planning algorithm for a 7-DOF robot arm operating in tight scenarios with sensor and motion uncertainties.
\end{abstract}

\input{1.tex}
\input{2.tex}
\input{3.tex}
\input{4.tex}
\input{5.tex}
\input{6.tex}

\bibliographystyle{IEEEtran}
\bibliography{IEEEabrv,refs}

\clearpage
\input{appendix.tex}

\end{document}

%% file: 1.tex
\section{Introduction}

Efficient collision detection is an important problem in robot motion planning, physics-based simulation, and geometric applications.
Earlier work in collision detection focused on fast algorithms for rigid convex polytopes and non-convex shapes and later extended to non-rigid models~\cite{Lin03collisionand,klosowski1998efficient}.
Most of these methods assume that an exact geometric representation of the objects is known in terms of triangles or continuous surfaces~\cite{gress2006gpu} and the output of collision query is a simple binary outcome.

As robots navigate and interact with real-world objects, we need algorithms for motion planning and collision detection that can handle environmental uncertainty. 
In particular, robots operate with sensor data, and it is hard to obtain an exact shape or pose of an object.
For example, depth cameras are widely used in robotics applications and the captured representations may have errors that correspond to lighting, calibration, or object surfaces~\cite{khoshelham2012accuracy}.
This gives rise to probabilistic collision detection, where the goal is to compute the probability of in-collision state by modeling the uncertainty using some probabilistic distribution.

Many collision detection algorithms have been proposed to account for such uncertainties~\cite{rusu2009real,bae2009closed,pan2011probabilistic}.
In practice, it is hard to analytically compute the collision probability for all probabilistic representations of uncertainties.
Most prior work on probabilistic collision detection is limited to Gaussian distributions~\cite{xu2014motion,park2017efficient}.
However, these formulations may not work well when objects are captured by a depth sensor.
Even the process of capturing static objects can result in depth images, where the depth values can vary between consecutive frames due to different noise sources.
The dynamic objects in the scene can pose additional problems due to sensor noise and the uncertainties introduced by the objects' motions.
Moreover, Gaussian process dynamical models used to represent human motion~\cite{wang2008gaussian} have an inherited uncertainty in the Gaussian variances as the central motion is represented using  Gaussian means.

\begin{figure}[t]
  \centering
  \subfloat
  {
    \includegraphics[width=0.31\linewidth]{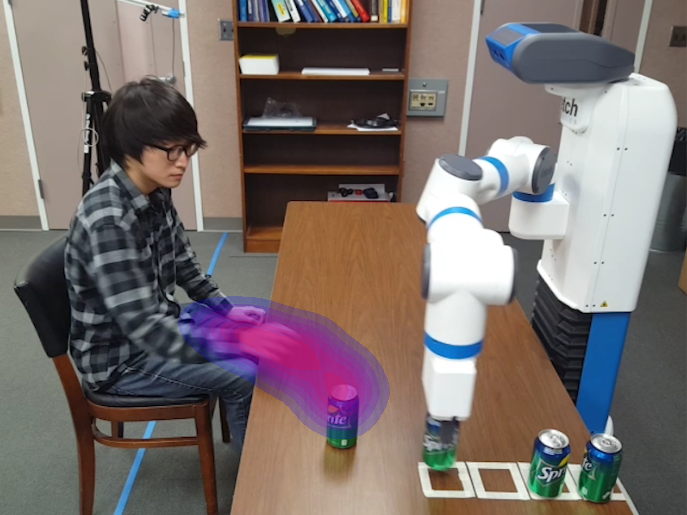}
  }
  \subfloat
  {
    \includegraphics[width=0.31\linewidth]{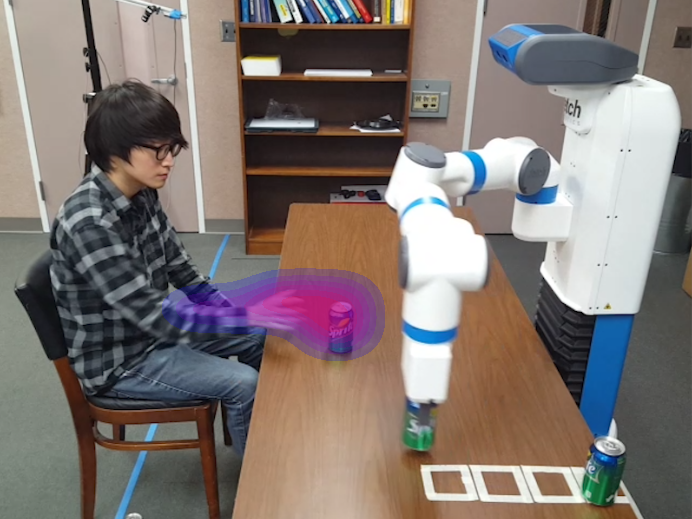}
  }
  \subfloat
  {
    \includegraphics[width=0.31\linewidth]{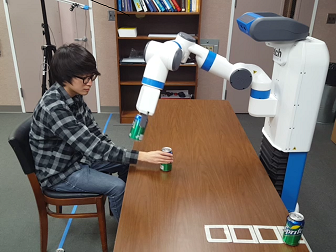}
  }
  \caption{We highlight the benefits of our novel probabilistic collision detection with a Truncated Gaussian error distribution. Our formulation is used to accurately predict the future human motion and integrated with a motion planner for the 7-DOF Fetch robot arm. As compared to prior probabilistic collision detection algorithms based on Gaussian distribution~\cite{park2017efficient}, our new method improves the running time by $2.6$x and improves the accuracy of collision detection by $9.7$x.
  }
  \label{fig:pcd_real_robot}
  \vspace*{-0.15in}
\end{figure}

In many applications, it is necessary to use non-Gaussian models for uncertainties~\cite{kurniawati2013online}.
These include Truncated Gaussian with bounded domains for sensory noises~\cite{johnson1994lognormal,simon2006optimal} to represent the position uncertainties for a point robot position~\cite{patil2012estimating}.
Other techniques model the uncertainty as a Partially Observable Markov Decision Process (POMDP)~\cite{rafieisakhaei2016non,seiler2015online}.

\noindent {\bf Main Results:}
We present efficient algorithms to compute the collision probability for error distributions corresponding to a variety of non-Gaussian models, including weighted samples and Truncated Gaussian (TG).
Our approach is based on modeling the TG error distribution and represent the collision probability using a volume integral (Section 4).
We present efficient techniques to evaluate the integral and highlight the benefits over prior methods for probabilistic collision detection.
We evaluate their performance on synthetic as well as real-world datasets captured using depth cameras (Section 5). Furthermore, we show that our efficient probabilistic collision detection algorithm can be used for real-time robot motion planning of a 7-DOF manipulator in tight scenarios with depth sensors. Some novel components of our work include:

\begin{itemize}
\item A novel method to perform probabilistic collision detection for TG Mixture Models based on appropriately formulating of the vector field, and computing an upper bound using divergence theorem on the resulting integral.
Moreover, we present an efficient method to evaluate this bound for convex and non-convex shapes.
\item 
We show that TG outperforms normal Gaussian, and Truncated Gaussian Mixture Model (TGMM) outperforms Gaussian Mixture Model (GMM). In practice, probabilistic formulation is less conservative than prior methods and results in $5-9\times$ accuracy in terms of collision probability computation (Table 1).
\item We have combined our probabilistic collision formulation with an optimization-based realtime robot motion planner that accounts for positional uncertainty from depth sensors.
Our modified planner is less conservative in terms of computing paths in tight scenarios.
\end{itemize}

%% file: 2.tex
\section{Related Work}

We give a brief overview of prior work on probabilistic collision detection.

\subsection{Probabilistic Collision Detection for Gaussian Errors}

Many approaches to compute the collision probability in uncertain robotic environments approximate the noises using a single Gaussian or a mixture of Gaussian distributions to simplify the computations.
Such approaches are widely used in 2D environments for autonomous driving cars to avoid collisions with cars or pedestrians.
Xu et al.~\cite{xu2014motion} use Linear-Quadratic Gaussian to model the stochastic states of car positions on the road. Collision detection under uncertainty is performed by computing the Minkowski sum of Gaussian ellipse boundary and the rectangular car model and checking for overlap with other rectangular car model.
Park et al.~\cite{park2017efficient} present An efficient algorithm to compute upper bound of collision probability with Gaussian error distributions ~\cite{park2017efficient}. 
This approach can be extended to Truncated Gaussian because the probability density function (PDF) of a Truncated Gaussian inside its ellipsoidal domain has the same value  as that of the PDF of a Gaussian. Therefore, the  upper bound computed using ~\cite{park2017efficient}  also holds for Truncated Gaussian error distributions, but the bound is not tight.
Moreover, a Truncated Gaussian distribution has a bounded ellipsoidal domain and the integral computations outside the domain can be omitted. As compared to this approach, our new algorithm improves the tightness of the upper bound and the running time, as shown in Section 5.

\subsection{Probabilistic Collision Detection for Non-Gaussian Errors}

The collision probability for non-Gaussian error distributions can be computed with Monte Carlo sampling~\cite{du2011probabilistic}. However, these methods are much slower  ($10-1000$ times), as compared to probabilistic algorithms that use  Gaussian forms of error distributions~\cite{park2017efficient}.
Althoff et al.~\cite{althoff2009model} use a non-Gaussian probability distribution model on the future states of other cars on the road, based on their positions, speeds, and road geometry. They use a 2D grid discretization of the state space and  Markov chain to compute the probability that a car belongs to a cell. This method assumes that the environment sensors has no noise.
Lambert et al.~\cite{lambert2008fast} use a Monte Carlo approach, taking advantage of the probability density function represented as a Gaussian. Other methods have been proposed for point clouds using classification~\cite{pan2011probabilistic} or Monte-Carlo integration~\cite{park2016fast}.
Approaches based on Partially Observable Markov Decision Processes (POMDPs) make efficient decisions about the robot actions in a partially observable state in an uncertain environment~\cite{kurniawati2013online,bai2015intention}.
Some applications using POMDPs~\cite{van2012lqg} have been developed to avoid collisions in an uncertain environment, where the uncertainty is represented with a non-Gaussian probability distribution. Our approach for non-Gaussian distributions is different and complementary with respect to these methods.

\subsection{Probabilistic Collision Detection: Applications}

Many approaches have been proposed for collision checking for general applications.
Aoude et al.~\cite{aoude2013probabilistically} represent the uncertainty model for point obstacles as a Gaussian Process and positional error is represented by a Gaussian distribution that propagates over a discretized time domain. The upper bound on collision probability is computed on the Gaussian positional error with an $\text{erf}(\cdot)$ function for a point obstacle.
Fisac et al.~\cite{fisac2018probabilistically} compute the collision probability between the dynamic human motion and a robot, and use that value for robot motion planning in the 3D workspace. This algorithm models the human motion based on human dynamics, discretizes the 3D workspace into smaller grids, and integrates the cell probabilities over the volume occupied by the robot.
Probabilistic collision detection for a Gaussian error distribution~\cite{park2017efficient} has been used for optimization-based robot motion planning.
The collision constraint used in the optimization formulation is that the collision probability should be less than $5\%$ at any robot configuration in the resulting trajectory.
However, with Gaussian error distributions, the upper bound of collision probability is rather conservative. As a result, these approaches do not work well in tight spaces or narrow passages.

%% file: 3.tex
\section{Overview}

In this section, we introduce the terminology used in the paper and give an overview of our approach. Our algorithm is designed for environments, where the 
 scene data is captured using sensors and only partial observations are  available.
In this case, the goal is to compute the {\em collision probability} between two objects, when one or both objects are represented with uncertainties and some of the input information such as positions or orientations of polygons or point clouds are given as probability distributions

\subsection{Probabilistic Collision Detection}

The input of the probabilistic collision detection is two 3D shapes $A$ and $B$, and  two 3D positional error distributions $P_A$ and $P_B$ that are probabilistically independent of each other.
The positional error distributions $P_A$ and $P_B$ denote the probability density function over 3D space of translations from the origins of objects $A$ and $B$, respectively.
The output of the algorithm is $p_{col}$, the probability of in-collision state between $A$ and $B$, where the objects can be translated with the error distributions.

The collision probability $p_{col}$, given two input shapes $A$ and $B$ and the error distributions $p_A$ and $p_B$, can be formulated as
\begin{align}
    p_{col} = \iiint_{\epsilon_A} \iiint_{\epsilon_B} & I \left( (A + \epsilon_A) \cap (B + \epsilon_B) \neq \varnothing \right) \nonumber \\
    & p(\epsilon_A) p(\epsilon_B) d\epsilon_A d\epsilon_B,
\end{align}
\begin{align}
    \epsilon_A \sim P_A, \quad \epsilon_B \sim P_B,
\end{align}
where $I(\cdot)$ is an indicator function which yields 1 if the condition is true and 0 otherwise, $\epsilon_A$ and $\epsilon_B$ are the displacement vectors for $A$ and $B$ with the probability distribution $P_A$ and $P_B$, and $\bigoplus$ denotes the Minkowski sum operator between two shapes.

To generalize, we shift only one object $A$ by $\epsilon = \epsilon_A - \epsilon_B$ which follows a probabilistic distribution $P_{AB}$, instead of shifting the two objects separately by $\epsilon_A$ and $\epsilon_B$.
Because of the independence of probabilistic distributions $P_A$ and $P_B$, the convolution $P_{AB}$ of $P_A$ and $P_B$ can be expressed as:
\begin{align}
    f_{AB}(\mathbf{x}) = \iiint_{\mathbf{y}} f_{A}(\mathbf{y}) f_{B}(\mathbf{x}-\mathbf{y}) d\mathbf{y}, \label{eq:convolution}
\end{align}
where $f_{AB}$, $f_A$, $f_B$ are the probability density functions of $P_{AB}$, $P_A$, $P_B$, respectively.

\subsection{Probabilistic Collision Detection for Gaussian Error}

The general probabilistic collision detection problem is hard to solve, when the error distributions $P_A$ and $P_B$ have any arbitrary form.   
The convolution operator in (Equation (\ref{eq:convolution})) can be hard to formulate in the general case. 
However, it is known that the convolution of two Gaussians is also Gaussian.
This generalizes the use of two error distributions into one, yielding the following:
\begin{align}
    p_{col} &= \iiint_{\epsilon} I \left( (A + \epsilon) \bigoplus B) \neq \varnothing \right) p(\epsilon) d\epsilon \label{eq:cp1} \\
    &= \iiint_{\epsilon} I \left( \epsilon \in (-A) \bigoplus B \right) p(\epsilon) d\epsilon, \quad \epsilon \sim P_{AB}. \label{eq:cp2}
\end{align}
Probabilistic collision detection with the Gaussian distribution condition can be solved efficiently~\cite{park2017efficient}, where $P_A$ and $P_B$ also correspond to Gaussian distributions. This algorithm computes a good upper bound on collision probability for convex and non-convex shapes by efficiently linearizing the Gaussian along the minimum displacement vector direction. In practice, the resulting bounds are conservative.

%% file: 4.tex
\section{Truncated Gaussian Mixture Model Error Distribution}

In this section, we present an efficient algorithm for
Truncated Gaussian Mixture Model (TGMM) error distributions, which is a more general type of noise model for robotics applications.
To compute the collision probability for TGMM, we first introduce the solutions for simpler error distributions corresponding to Truncated Gaussian (TG) and Weighted Samples (WS). We combine these two algorithms to design an algorithm for a multiple Truncated Gaussian error distribution model.

\subsection{Truncated Gaussian Mixture Models}

A TGMM consists of multiple Truncated Gaussian (TG) distributions, each distribution with a truncated domain.
The probability density function of a TG, $f_{TG}$, can be formulated as:
\begin{align}
    f_{TG}(\mathbf{x}; \mu, \Sigma, r) &=
    \begin{cases}
        \frac{1}{\eta} g(\mathbf{x} ; \mu, \Sigma) & (\mathbf{x} - \mu)^T \Sigma^{-1} (\mathbf{x} - \mu) \leq r \\
        0 & \text{otherwise}
    \end{cases}, \label{eq:pdf_tg}
\end{align}
where $g$ is the probability density function of a Gaussian, $\mu$ is the mean, $\Sigma$ is the variance, $r$ is the radius of bound in the coordinates of the principal axes, and $\eta$ is the truncation rate used to compensate the loss of truncated volume of probability outside the bound.
A TGMM consists of $n$ TGs with multiple weights $w_i$.
The probability density function of the TGMM, $f_{TGMM}$, can be formulated using the definition of $f_{TG}$ in Equation (\ref{eq:pdf_tg}), as:
\begin{align}
    f_{TGMM}(\mathbf{x}) = \sum_{i=1}^{n} w_i f_{TG}(\mathbf{x}; \mu_i, \Sigma_i, r_i), \quad \sum_{i=1}^n w_i = 1.
\end{align}
As the radii of TGs decrease and converge to zero, the probability model behaves like a discrete probability distribution, which we call Weighted Samples (WS).
The WS is a discrete probability distribution, formulated as:
\begin{align}
    P(\mathbf{X} = \mathbf{x}_i) = w_i, \quad \sum_{i=1}^n w_i = 1 \label{eq:pdf_ws}
\end{align}
where $\mathbf{x}_i$ is a sample in ${\rm I\!R}^d$, and $w_i$ is a weight of the sample, for $i=1,\cdots,n$.

\subsection{Collision Probability for Truncated Gaussian}

The TG is a Gaussian with a specific form of bounded domain. The bounded domain for 3D Truncated Gaussian is an ellipsoid, centered at the Gaussian mean and having the same principal axes as those of Gaussian variances.
The TG is formulated with a collision probability function as
\begin{align}
    p_{col} &= \iiint_{V_{AB}} f_{TG}(\mathbf{x}; \mu, \Sigma, r) d\mathbf{x},
\end{align}
where $V_{AB} = -A \bigoplus B$, $f_{TG}$ is the probability density function for TG, $\mu$ is the mean, $\Sigma$ is the variance, $r$ is the radius of bound in the coordinates of the principal axes, and $\eta$ is the normalization constant used to compensate the loss of truncated volume of probability outside the bound.
Because $f_{TG}$ has the value of a Gaussian multiplied by $\eta$ inside the boundary, the integral volume becomes $-A \bigoplus B \cap V_{TG}$, where $V_{TG}$ is the valid volume of Truncated Gaussian.
The collision probability corresponds to
\begin{align}
    p_{col} = \frac{1}{\eta} \iiint_{V_{AB} \cap V_{TG}} f_{TG}(\mathbf{x}; \mu, \Sigma, r) d\mathbf{x}. \label{eq:cp_tg}
\end{align}
The TG has its center at $\mu$ and principal axes with different lengths determined by $\Sigma$.
To normalize the function, a transformation $T = \Sigma^{-1/2} - \mu I$ is applied to the coordinate system, which changes Equation (\ref{eq:cp_tg}) to
\begin{align}
    p_{col} = \frac{1}{\eta \det \Sigma} \iiint_{V'_{AB} \cap V'_{TG}} f_{TG}(\mathbf{x}; \mathbf{0}, I, r) d\mathbf{x}, \\
    V'_{AB} = T(V_{AB}), \quad V'_{TG} = T(V_{TG}).
\end{align}
In the transformed coordinate system, $V'_{TG}$ is a sphere of radius $r$.

Unfortunately, there is no explicit or analytic form of solution for the integral of a Gaussian distribution over the intersection of a non-convex volume $V'_{AB}$ and a ball $V'_{TG}$. In order to simplify the problem, we initially assume that $A$ and $B$ are convex, and so are $V_{AB}$ and $V'_{AB}$.
Instead of computing the exact integral, we compute an upper bound on the collision probability.
The computation of collision probability reduces to the computation of the integral
\begin{align}
    \iiint_{V'} g(\mathbf{x}; \mathbf{0}, I) d\mathbf{x},
\end{align}
where $V' = V'_{AB} \cap V'_{TG}$, and $g(\cdot)$ is the Gaussian probability density function.

From the convexity of $V'_{AB}$, the minimum distance vector $\mathbf{d}'$ between the origin and $V'_{AB}$ can be computed by using the GJK algorithm~\cite{gilbert1988fast} between $A'$ and $B'$, which are transformed from $A$ and $B$ by $T$.
Let $\mathbf{n}_d'$ be the unit directional vector of $\mathbf{d}$'.
Then, by the Cauchy-Schwarz inequality $(\mathbf{x} \cdot \mathbf{n}_d')^2 \leq \| \mathbf{x} \|^2$, the integral is bounded by
\begin{align}
    p_{col} \leq \iiint_{V'} \frac{1}{\sqrt{8 \pi^3}} \exp \left( -\frac{1}{2} (\mathbf{x} \cdot \mathbf{n}_d')^2 \right) d \mathbf{x} . \label{eq:pdf_upper}
\end{align}
The integrand of the upper bound term behaves as a 1D Gaussian function instead of being the 3D function. We use the divergence theorem to compute the upper bound on collision probability (\ref{eq:pdf_upper}).
\begin{align}
    \iiint_{V'} \text{div} (\mathbf{F}) dV = \oiint_{S'} (\mathbf{F} \cdot \mathbf{n}_S) dS , \label{eq:div_equality}
\end{align}
where $\mathbf{F}$ is a vector field, $S'$ is the surface of $V'$, $dS$ is an infinitesimal area for integration, and $\mathbf{n}_S$ is the normal vector of $dS$.
This converts the volume integral to a surface integral.
Let's define $\mathbf{F}$ as
\begin{align}
    \mathbf{F}(\mathbf{x}) = \frac{1}{2 \pi} \left( 1 + \text{erf} \left( \frac{\mathbf{x} \cdot \mathbf{n}'_d}{\sqrt{2}} \right) \right) \mathbf{n}'_d ,
\end{align}
where $\text{erf}(\cdot)$ is the 1D Gaussian error function.
Note that $F$ is a vector field with a single direction $\mathbf{n}'_d$.
The directional derivative of $\mathbf{F}(\mathbf{x})$ along any directional vector orthogonal to $\mathbf{n}_d'$ is zero because $\mathbf{F}$ varies only along $\mathbf{n}_d'$.
The divergence of $\mathbf{F}$ thus becomes $(\partial \mathbf{F} / \partial \mathbf{n}_d')$, and this is equal to the function in Equation (\ref{eq:pdf_upper}).

We apply the divergence theorem in Eqation (\ref{eq:div_equality}) to the volume integral on $V'$ in Equation (\ref{eq:pdf_upper}).
Note that $V'$ is a 3D volume intersection between a non-convex polytope $V'_{AB}$ and a ball $V'_{TG}$.
The surface integral on the intersection between a non-convex polytope and a ball can be decomposed into two parts and bounded by the sum of two components as
\begin{align}
    \sum_{i} \oiint_{\triangle S'_{i}} (\mathbf{F} \cdot \mathbf{n}'_i) dS + \oiint_{S'_{TG}} (\mathbf{F} \cdot \mathbf{n}_S) dS,
\end{align}
where $S'_{i}$ is the $i$-th triangle of $V'_{AB}$ inside $V'_{TG}$, $\mathbf{n}'_i$ is the normal vector of $\triangle S'_{i}$, and $S'_{TG}$ is the spherical boundary of $V'_{TG}$ outside of a plane defined by $\mathbf{d}'$.
The second term corresponds to the spherical domain of the normalized Truncated Gaussian with the truncation rate $\eta$.
The magnitude of $F$ on the spherical boundary $V'_{TG}$ is upperly bounded by $(1 - \eta)$, because it is the cumulative distribution function on the boundary.
The surface area of $S'_{TG}$ is less than $\pi ||\mathbf{d}'||^2$.
This can be used to express a bound based on the following lemma.
\begin{lemma}
{\em The collision probability represented in a volume integral is upperly bounded by a surface integral as follows:
\begin{align}
    p_{col} &= \iiint_{V'} g(\mathbf{x}; \mathbf{0}, I) d\mathbf{x} \\
    &\leq \sum_{i} \oiint_{\triangle S_{i}} (\mathbf{F} \cdot \mathbf{n}_i) dS + \pi (1 - \eta) ||\mathbf{d}'||^2,
\end{align}
where $F$ is a vector field in 3D space whose maximum magnitude is $1/\pi$, and $S_{i}$ is the $i$-th triangle of $V'_{AB}$ that is inside $V'_{TG}$.}
\end{lemma}

Because the error function integral over a triangle domain is hard to compute, the upper bound on the integral is evaluated as
\begin{align}
    & \sum_{i} \oiint_{\triangle S_{i}} (\mathbf{F} \cdot \mathbf{n}_i) dS \\
    \leq & \sum_i \left( \max_{j=1,2,3} \mathbf{F}(S_{ij}) \cdot \mathbf{n}_i \right) \text{Area}(\triangle S_{i}), \label{eq:upper_bound}
\end{align}
where $S_{ij}$ is the $j$-th vertex of the triangle $S_{i}$ for $j \in \{1,2,3\}$.
The upper bound on the collision probability corresponds to the sum of the maximum of $F$ at the points of each triangle, multiplied by the area of the triangle, over the surface of $V'_{AB} \cap V'_{TG}$.

\begin{figure}[t]
  \centering
  \subfloat[][]
  {
    \includegraphics[width=0.33\linewidth]{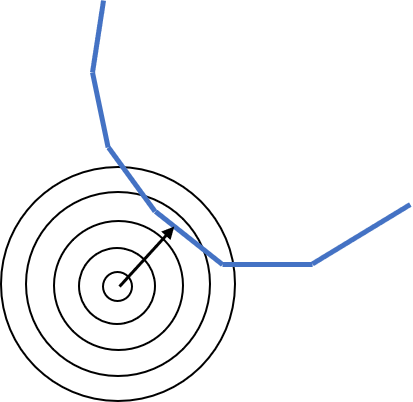}
  }
  \subfloat[][]
  {
    \includegraphics[width=0.33\linewidth]{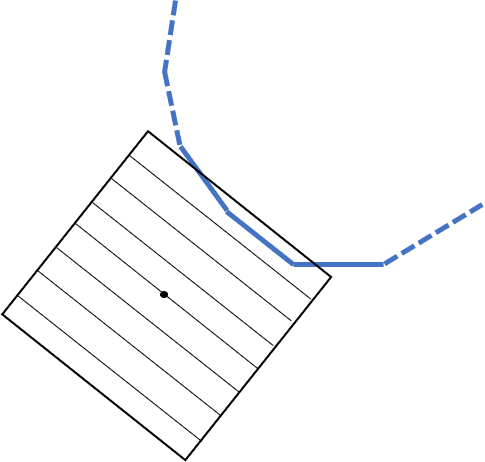}
  }
  \caption{(a) Contour plots of the bivariate TG distribution. (b) Contour plots of the bounded function $\mathbf{F}$ for TG are not used in the calculation of collision probability and thereby reduce the running time of collision probability computation.}
  \label{fig:truncated_pdf}
  \vspace*{-0.2in}
\end{figure}

\begin{figure}[t]
  \centering
  \includegraphics[width=0.75\linewidth]{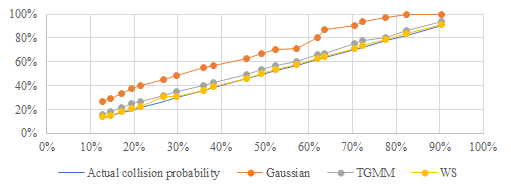}
  \caption{The upper bound of collision probability with uncertainty approximated as Gaussian, Gaussian Mixture, and Weight Samples. The X-axis is the true collision probability computed using Monte Carlo methods, and Y-axis is the computed probability using different methods. The computed upper bound for Gaussian Mixture and Weights Samples are closer to the ground truth/exact answer, than that for a single Gaussian approximation.
  The collision probability over-estimation with TGMM is reduced by 90\%, compared to the one with Gaussian distribution.
  }
  \label{fig:pcd_result1}
  \vspace*{-0.2in}
\end{figure}

\subsection{Efficient Evaluation of the Integral}
In order to reduce the running time of computing the surface integral, we take advantage of the bound of TG.
The domain of surface integral is $V' = V'_{AB} \cap V'_{TG}$, where $V'_{AB}$ consists of many triangles and $V'_{TG}$ is a sphere of radius $r$.
This sphere is tightly bounded by a cube, with one normal parallel to the direction of shortest displacement vector $\mathbf{d}'$.
Therefore, the triangles of $V'_{AB}$ that are outside the cube do not count towards the surface integral.
So, we accumulate the upper bound function value in Equation (\ref{eq:upper_bound}) only for the triangles that lie inside the cube boundary, and ignore the triangles outside the boundary.
For the triangles that intersect the cube boundary, the upper bound function value is computed for the intersecting primitives.
Because the approximated integral for collision probability is bounded by the cube, primitives outside the cube can be ignored in terms of calculating the upper bound of collision probability.
Limiting the computation to the truncated primitives can accelerate the running time. 

In order to perform this computation for non-convex primitives, we construct bounding volume hierarchies (BVHs) for $A$ and $B$, with each bounding volume being an oriented bounding box.
During the traversal of the BVHs, the oriented bounding boxes are first transformed by $T$.
The transformed bounding volumes are still convex primitives, the surface integral can be obtained using Equation (\ref{eq:upper_bound}).

\subsection{Error Distribution as Weighted Samples}

For the weighted samples, the probability distributions are given by multiple points $p_i$ with weights $w_i$, yielding a discrete probability distribution, as described in Equation (\ref{eq:pdf_ws}).
The collision probability of Equation (\ref{eq:cp1}-\ref{eq:cp2}) for the weighted samples is given as:
\begin{align}
    p_{col} = \sum_{i=1}^n w_i I \left( \mathbf{x}_i \in (-A) \bigoplus B \right), \quad \sum_{i=1}^n w_i = 1,
\end{align}
where $w_i$ is weight and $I(\cdot)$ is the indicator function which yields $1$ if the statement inside is true or $0$ otherwise.
The formulation is the weighted average of $n$ collision detection results.
A simple solution to this problem is to run exact collision detection algorithms $n$ times and sum up the weights of in-collision cases. However, this results in an $O(n)$ and we use BVHs to accelerate that computation.

We have the bounding volumes for the weighted samples and the two polyhedra.
When there is no overlap between the bounding boxes, it implies that there is no collisions between two shapes for all weighted samples in the corresponding bounding volume.
If the bounding volumes overlap, there may be a collision for each weighted samples, and the bounding volumes of the children are checked recursively for collisions. Each of these bounding volume checks can be performed in $O(1)$ time.

If we want to compute an upper bound of collision probability, the running time can be further reduced by replacing detailed computation of collision probability with a simple upper bound.
We introduce a user-defined parameter $\delta$ which we call the ``confidence level''.
During the traversal of bounding volume traversal tree, the upper bound of collision is the sum of weights of samples that belong to the bounding volume.
If the upper bound is less than the confidence level $1 - \delta$, the traversal stops and the sub-routine returns the sum of weights as an upper bound of collision probability.

In order to reduce the time complexity for more complex forms of error distributions, we construct a Bounding Volume Hierarchy (BVH)~\cite{Lin03collisionand} over the error distributions of mixture models with Oriented Bounding Boxes (OBBs)~\cite{gottschalk1996obbtree} and apply the collision probability algorithm on its nodes, which are convex primitives.
We construct a BVH for the weighted samples and for Truncated Guassian Mixture Models in $O(n)$ time complexity.
The BVH is generated from the root node that contains every Truncated Gaussians, and the bounding volume for the root node is computed by minimizing the volume of the oriented bounding box.
Next, the bounding volume is split at the center along the longest edge and two child BVH nodes are generated, each containing appropriate samples. This process is repeated till the leaf nodes.

\subsection{Error Distribution as Truncated Gaussian Mixture Models}
For TGMM the probability distribution is given as:
\begin{align}
    p_{col} = \sum_{i=1}^n w_i \iiint_{V_{AB}} \eta_i f_{TGMM}(\mathbf{x} ; \mu_i, \Sigma_i, r_i) d\mathbf{x},
\end{align}
where $n$ is the number of Truncated Gaussians (TGs), $w_i$ is the weight of each TG, and $\mu_i$, $\Sigma_i$ and $r_i$ are the mean, variance and radius of TGs, respectively. The overall algorithm for  TGMM is obtained by combining the two previous algorithms.
A change from the algorithm for weighted samples is that the BVH is constructed for $n$ TGs with their ellipsoid bounds instead of the point samples.
The details of algorithms and pseudo-codes are given in the appendix~\cite{1902.10252}.

%% file: 5.tex
\section{Performance and Analysis}

In this section, we describe our implementation and  highlight the performance of our probabilistic collision detection algorithms on synthetic and real-world benchmarks.
Furthermore, we measure the upper bound of collision probabilities and speedups for algorithms with different noise distributions, compared to the exact collision probability computed by Monte Carlo method.
\subsection{Probabilistic Collision Detection: Performance}

\begin{figure}[t]
  \centering
  \subfloat[][]
  {
    \includegraphics[width=0.75\linewidth]{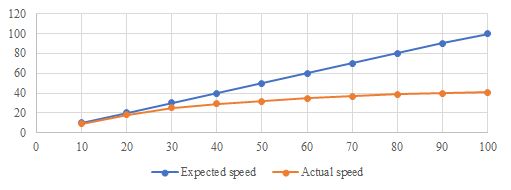}
  }
  \\
  \subfloat[][]
  {
    \includegraphics[width=0.75\linewidth]{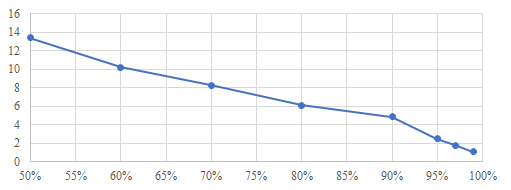}
  }
  \caption{
  (a) Speedup of weighted samples (expected) case compared to Monte Carlo (actual) with between 10 to 100 samples (X-axis).
  (b) Speedup of Truncated Gaussian case, compared to the running time of probabilistic collision detection with a Gaussian.
  X-axis is the untruncated volume of Gaussian, meaning 100\% is the Gaussian and lower value indicates smaller bound.
  As the truncation boundary shrinks up to 50\% of the volume of Gaussian, the algorithm with TG is  $14$x faster times than the algorithm Gaussian distribution.
  }
  \label{fig:pcd_result2}
  \vspace*{-0.15in}
\end{figure}

For the translational error distribution, we first generate a ground truth distribution by randomizing the parameters of TGMM.
Next, we sample 100,000 points from the distribution.
We run expectation-maximization from these samples to find parameters of single Gaussian, single TG, WS, and TGMMl.
The resultant distributions are different from the ground truth and count toward collision probability over-estimation.

Figure~\ref{fig:pcd_result1} shows the collision probabilities of noise models approximated with Gaussian, TGMM of $10$ TG distributions, and $100$ WS.
We observe that the  approximation with a single Gaussian yields rather high and conservative value  of collision probability, compared to the ground truth collision probability.
Figure~\ref{fig:pcd_result2} (a) shows the speedup of probabilistic collision detection with WS (Algorithm 2 in the Appendix) over the Monte Carlo method.
The collision probability computation with Monte Carlo counts the sum of sample weights for every Weighted Sample that is in collision,  and is similar to an exact collision detection algorithm.
Thus, the running time of Monte Carlo increases linearly as the number of Weighted Samples increases.
Figure~\ref{fig:pcd_result2} (b) shows the speedup of probabilistic collision detection with a TG model (Algorithm 1 in the Appendix) compared to probabilistic collision detection with a Gaussian error distribution.
In case of TGs, BVH traversal is not performed when the truncation boundary of TGs does not overlap with the Minkowski sum of bounding volumes for the two objects. 
On the other hand, for Gaussian distribution there are no truncation boundaries and the BVH traversal continues. Thereore, we observe speedup with TGs over Gaussian, as shown in Figure~1 and Table 1.
Overall, the speedup depends on the range of truncation. A smaller truncation boundary results in faster  performance of our probabilistic collision detection algorithm.

\subsection{Sensor Noise Models for Static Obstacles}

\begin{figure}[t]
  \centering
  \subfloat[][]
  {
    \includegraphics[width=0.3\linewidth]{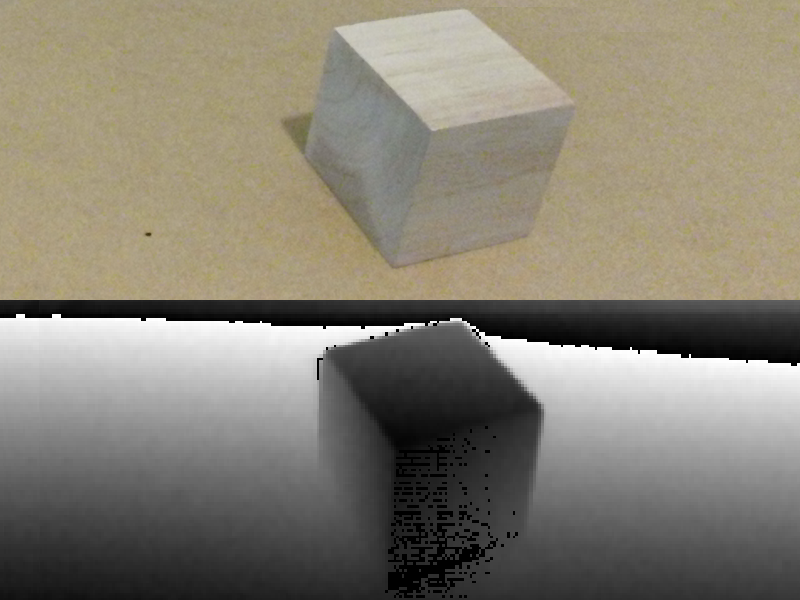}
  }
  \subfloat[][]
  {
    \includegraphics[width=0.3\linewidth]{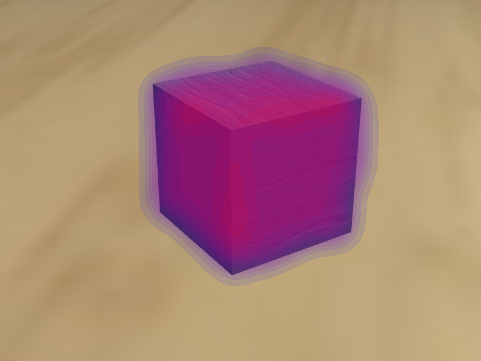}
  }
  \subfloat[][]
  {
    \includegraphics[width=0.3\linewidth]{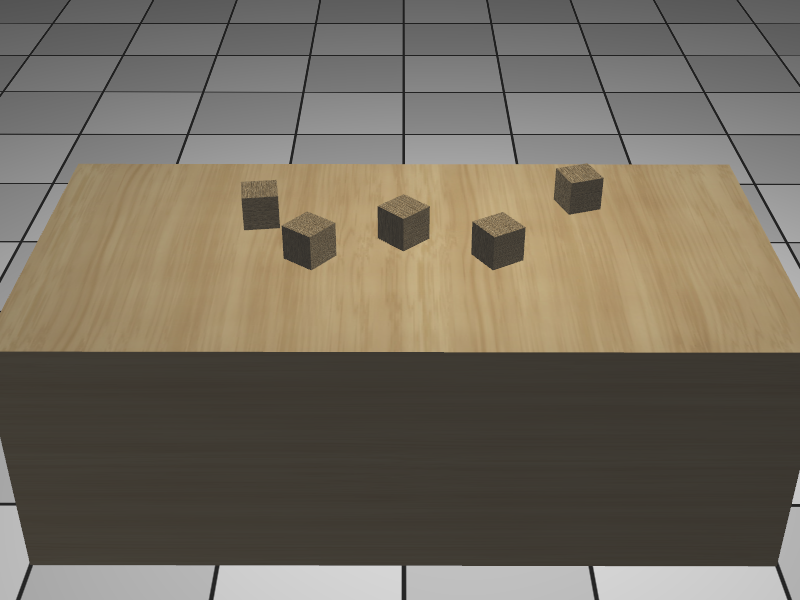}
  }
  \caption{(a) A captured RGBD image. The depth values of the table and the wood block have noises, even in adjacent frames. The TG noise of each point particle of the wood block contributes to the overall TGMM model.  (b) A reconstructed 3D model of a wood block with TGMM, which bounded around the wooden block and more acuracy than Gaussian distribution, which has unbounded probability density function.
  (c) A reconstructed 3D robot environment with error distributions on the table and the wood blocks. The wood blocks placed in a zig-zag pattern result in 4 narrow passages for the robot. 
  }
  \vspace*{-0.20in}
  \label{fig:pcd_static_noise}
\end{figure}

In a real-world setting, we add a noise model to the point cloud data.
The variance is chosen based on the Kinect sensor uncertainty.
The input depth images have noise in each pixel and, according to~\cite{khoshelham2012accuracy}, the noise of each pixel can be approximated with a 1D Gaussian.
Thus, noise in the pixels of an object are combined with a Gaussian Mixture noise model for the object.
Depth images of the wooden blocks and the table have noise in each pixel.
We capture sequences of the depth images.
In the experiments, the approximate poses of tables and wooden blocks are known a priori.
On the boundary of the wooden blocks, some pixels are always classified as a wooden block, some other pixels are classified as a wooden block or as background in different frames, and other pixels are always classified as background.
From these boundary pixels, the variance in x- and y-axis noise on the Kinect sensor coordinate frame can be set to the thickness of the boundary.
The variance in z-axis motion is computed from the always-wooden-block pixels.
The depth values on those pixels differ from frame to frame.
After the variance of noise Gaussian is calculated, the mean and variance of a Truncated Gaussian are the same.
The truncation rate $\eta$ with which the integral in the ellipsoidal boundary is set to 90\% in our benchmarks.
With the fixed truncation rate, the positional error is bounded around the wooden blocks, unlike the positional error represented by Gaussians with an unbound domain.
The confidence level $\delta$ is related to the robot motion planner, constraining that the collision probability between the robot and the objects should be less than $1 - \delta$ at any robot trajectory point.
In our benchmarks, $\delta$ is set to 95\%.

Figure~\ref{fig:pcd_static_noise} shows a captured depth image, a noise distribution modeled using Truncated Gaussian Mixture Model, and a reconstructed 3D environment with noises.
The pixels on the boundary of the object have higher variance in terms of noise.
So, the Truncated Gaussian Mixture noise model may have some Gaussians with higher variance.
A principal axis for those boundary pixels is perpendicular to the boundary direction.

\subsection{Robot Motion Planning}

\begin{table*}[ht]
  \centering
  \begin{tabular}{|c|c|c|c|c|c|c|c|c|c|c|c|}
    \hline
    \multirow{2}{*}{Algorithm} & \multicolumn{3}{|c|}{\begin{tabular}{@{}c@{}}Collision Probability \\ Over-estimation (\%p) \end{tabular}} & \multicolumn{3}{|c|}{Running Time (ms)} & \multirow{2}{*}{\# Passages} & \multirow{2}{*}{Success Rate} & \multirow{2}{*}{Distance (cm)} \\ \cline{2-7}
    & Min & Max & Avg & Min & Max & Avg & & & \\ \hline
    CD-Obstacles & 0 & 0 & 0 (0)           & 2.8 & 8.0 & 3.5 (0.75) & 4/4 & 10/10 & 1.2 (0.09) \\ \hline
    CD-Points    & 0 & 100 & 23 (16)       & 2300 & 2800 & 2500 (130) & 1/4 &  2/10 & 13 (8.8) \\ \hline
    PCD-Gaussian~\cite{park2017efficient} & 8.6 & 35 & 15 (5.2)     & 15 & 100 & 27 (23) & 3/4 &  8/10 & 7.9 (2.6) \\ \hline
    PCD-TG       & 2.5 & 13 & 5.0 (3.3)    & {\bf 5.2} & {\bf 86} & {\bf 12 (4.6)} & 3/4 &  7/10 & 6.0 (1.3) \\ \hline
    PCD-WS       & 0.72 & 8.1 & 1.5 (0.60) & 130 & 220 & 180 (32) & {\bf 4/4} & {\bf 9/10} & 4.5 (0.66) \\ \hline
    PCD-GMM & 1.9 & 7.7 & 6.2 (3.3) & 85 & 480 & 240 (55) & 3/4 & 8/10 & 7.1 (1.0) \\ \hline
    PCD-TGMM     & {\bf 0.26} & {\bf 3.3} & {\bf 0.8 (0.32)} & 35 & 130 & 97 (17) & {\bf 4/4} & {\bf 9/10} & {\bf 3.7 (0.46)} \\ \hline
  \end{tabular}
  \caption{{\bf Performance of probabilistic collision detection algorithms:} Evaluated as part of a motion planner with sensor data. The collision probability over-estimation is shown as percent point (\%p) with the minimum and maximum over-estimation. The values corresponds to the average over the time of the robot trajectory with standard deviation in parenthesis.
  The best performance is obtained PCD-TGMM algorithm in terms of collision probability estimation (i.e. tight bounds), successful handling of narrow passages, computing collision-free trajectories, among different  algorithms.}
  \label{table:motion_planning}
  \vspace*{-0.255in}
\end{table*}

The probabilistic collision detection algorithm is used in optimization-based motion planning~\cite{Park:2012:ICAPS}.
We use a 7-DOF Fetch robot arm in the motion planning. 
We highlight its performance in terms of improved accuracy and faster running time in Figure~\ref{fig:pcd_real_robot}.
In the robot environment, there is a table in front of the robot and the wood blocks on the table are the static obstacles of the environment.
The wood blocks are placed in a manner that pairs of them result in a narrow passage for the robot's end-effector.
The environment is captured using two depth sensors.
One sensor is Primesense Carmine 1.09 sensor, the robot head camera.
Another is Microsoft Kinect 2.0 sensor installed in the opposite direction with respect to the robot.
The point clouds of the table and the wood blocks captured by the two sensors are used to reconstruct the environment. In this case, the reconstructed table surface and wood block obstacles have errors due to the noise in the depth sensors.
Figure~\ref{fig:pcd_static_noise} (c) shows the reconstructed environment from depths sensors and the error distributions around the obstacles.
The robot arm's task is to move a wood block, drawing a zig-zag pattern that passes through the narrow passages between the wood blocks.
The objective is to compute a robot trajectory that minimizes the distance between the robot's end-effector and the table, and not resulting in any collisions.
The following metrics are used to evaluate the performance:
\begin{itemize}
    \item Collision Probability Over-estimation: Because we compute the upper bounds of collision probability in our algorithm, we measure the extent of collision probability over-estimation, the gap between the upper bound of collision probability and the actual collision probability.
    \item Running Time: The running time of collision detection algorithm. This excludes the running time of motion planning algorithm.
    \item \# Passages: The successful number of passes the robot makes between wood blocks.
    \item Success Rate: The number of collision-free trajectories, out of the total number of trajectory executions. For each execution, the wood block's positions are set following the error distribution.
    \item Distance: The distance between the robot's end-effector and the table. 
    A lower value is better, as it implies that the robot can interact with the environment in close proximity.
\end{itemize}
The collision probability over-estimation, running time, distance values are measured for robot poses of every 1/30 seconds over the robot trajectories.

We compare the performances of 5 different collision detection algorithms: exact collision detection with static obstacles without environment uncertainties (CD-Obstacles), exact collision detection with point clouds (CD-Points), probabilistic collision detection with Gaussian errors (PCD-Gaussian)~\cite{park2017efficient}, PCD with Truncated Gaussian (PCD-TG), PCD with weighted samples (PCD-WS), and PCD with Truncated Gaussian Mixture Model (PCD-TGMM).
The weighted samples are drawn from the TGMMs. For each TG distribution, one sample is drawn from its center.
Other samples are drawn from three icosahedrons with the same centers and different radii by uniformly dividing the truncation radius.

\noindent {\bf Benefits of Truncated Gaussian:} Table~\ref{table:motion_planning} shows the results of robot motion planning in the various scenarios with different algorithms.
The robot motion planning without uncertainties (CD-Obstacles) operates perfectly.
However, under sensor uncertainties, the exact collision detection with the point clouds (CD-Points) works poorly. Due to the sensor uncertainties, a high error at a pixel affects the collision detection query accuracy and the performance in narrow passages and success rate. 
Our  probabilistic collision detection algorithms results in better performances than  exact collision detection algorithm under the environment with sensor uncertainties. 
Compared to PCD-Gaussian and PCD-GMM, our algorithms for  non-Gaussian distributions (PCD-WS, PCD-TGMM) demonstrate   better performances w.r.t. different metrics.  Figure~\ref{fig:pcd_robot} (appendix) highlights the results: using probabilistic collision detection with Gaussian error distribution (Figure~\ref{fig:pcd_robot} (a));  with TGMM error distribution (Figure~\ref{fig:pcd_robot} (b)).  CD-Obstacles and CD-Points correspond to the exact case, where the poses are known and there is no uncertainty (i.e. the ground truth). The robot's trajectory is shown in the video.

\noindent {\bf Human Motion Prediction:} 
We also evaluate our algorithm on scenarios with humans operating close to the robot, as shown in Fig. 1 and the video.
In order to handle the uncertainty of future human motion, we use probabilistic collision detection between the robot and the predicted future human pose. Improvement in the accuracy of motion prediction results in better trajectories in terms of being collision-free, smoother and being  able to handle tight scenarios~\cite{park2019planner}.  We highlight the benefits of our PCD-TGMM algorithm on the resulting trajectory computation in the video.

%% file: 6.tex
\section{Conclusion and Limitations}

We present efficient probabilistic collision detection algorithms for the following forms of non-Gaussian error distributions: Truncated Gaussian, Weighted Samples, and Truncated Gaussian Mixture Model.
Compared to the exact collision detection algorithm and prior probabilistic collision detection algorithms for Gaussian error distribution, our new method can compute a tighter upper bound of collision probability and improves the running time. We have integrated this algorithm with a motion planner and highlights its benefits in narrow passage scenarios with a 7-DOF robot arm. Our algorithm can be used to 
 model non-Gaussian error distributions from noisy depth sensors and predicted human motion models. 

Our approach has some limitations. The truncation on the Truncated Gaussians has a form of ellipsoid with the center and principal axes that is the same as the original Gaussian distribution.
The ellipsoidal shape of truncation boundary may not be sufficient for representing general error distributions. Another realistic possibility would be to truncate using planar boundaries.
In our future work, we would like to develop an algorithm for more generic non-Gaussian positional errors.
 We only consider the positional errors on obstacles and omit the rotational errors.
The rotational error cannot be approximated by Truncated Gaussian Mixture Model.
As part of future work, we would like to represent a rotational error distribution in the quaternion space, or in the affine space.
Furthermore, we assume that the noises of two objects A and B are independent, even though they may arise from the same source.

%% file: appendix.tex
\section*{Appendix}

We present some background on truncated Gaussian distributions as well as more details about our probabilistic collision detection algorithm  and its applications to motion planning. 

\subsection{Probabilistic Collision Detection}

We present the detailed pseudo-code of our novel probabilistic collision detection algorithms for Truncated Gaussian and weighted samples. 
\begin{algorithm}[h]
  \caption{$p_{col} = $PCD\_TG($T_A$, $T_B$, $\mu$, $\Sigma$, $r$, $\delta$) \\: Compute the collision probability between polyhedra $A$ and $B$, given precomputed BVHs for polyhedra $T_A$ and $T_B$, Truncated Gaussian parameters $\mu$, $\Sigma$, $r$, and a confidence level $\delta$.}
  \label{alg:pcd_tg}
  \begin{algorithmic}[1]
    \REQUIRE BVHs $T_A$ and $T_B$, a Truncated Gaussian $p(\cdot; \mu, \Sigma, r)$, and a confidence level $\delta$
    \ENSURE Upper bound on collision probability $p_{col}$
    \STATE $V_A$ = $T_A$.root
    \STATE $V_B$ = $T_B$.root
    \STATE $V'_{AB}$ = $T(-V_A \bigoplus V_B)$  \label{line:pcd_tg:minkowski_sum}
    \IF {IsNotCollision($V'_{AB}$, Sphere(r))}
      \RETURN $0$
    \ENDIF
    \STATE $\mathbf{d}'$ = GJK($T(V_A)$, $T(V_B)$)  \label{line:pcd_tg:gjk}
    \STATE $V'_{TG}$ = cube(norm($\mathbf{d}'$), $r$)  \label{line:pcd_tg:cube}
    \STATE $p_{col}$ := $0$
    \FORALL {$i$ of $V'_{AB}$}
      \IF {$\triangle S_{i1} S_{i2} S_{i3} \cap V'_{TG} \neq \varnothing$}  \label{line:pcd_tg:sum1}
        \STATE Add Equation (\ref{eq:upper_bound}) to $p_{col}$  \label{line:pcd_tg:sum2}
      \ENDIF
    \ENDFOR
    \IF {$p_{col} \leq \delta$ or both $T_A$ and $T_B$ are leaf nodes}  \label{line:pcd_tg:confidence1}
      \RETURN $p_{col}$  \label{line:pcd_tg:confidence2}
    \ENDIF
    \IF {$V_A$ has children and $Vol(V_A) \geq Vol(V_B)$}
      \RETURN $p_{col} = \sum\limits_{c_A : \text{child}}$PCD\_TG($c_A$, $T_B$, $\mu$, $\Sigma$, $r$, $\delta$)
    \ELSE
      \RETURN $p_{col} = \sum\limits_{c_B : \text{child}}$PCD\_TG($T_A$, $c_B$, $\mu$, $\Sigma$, $r$, $\delta$)
    \ENDIF
  \end{algorithmic}
\end{algorithm}

Algorithm~\ref{alg:pcd_tg} describes the overall process of computing the upper bound of collision probability for TG error distribution.
Line~\ref{line:pcd_tg:minkowski_sum} computes the Minkowski sum of two bounding volumes and transforms that sum using $T$.
Because the bounding volumes are oriented bounding boxes, the time complexity is $O(1)$
Line~\ref{line:pcd_tg:gjk} computes the minimum distance vector in nearly constant time and is used to define the function $F$ for Equation (\ref{eq:upper_bound}). The approximated cube boundary $V'_{TG}$ of TG is computed at Line~\ref{line:pcd_tg:cube}, whose center is at the origin and one of the normal directions is parallel to $\mathbf{d}'$, and whose half-length is $r$.
In Line~\ref{line:pcd_tg:sum1}-\ref{line:pcd_tg:sum2}, the upper bound of collision probability is computed.
If the computed collision probability bound is less than the confidence level, as shown in Line~\ref{line:pcd_tg:confidence1}, the algorithm returns the bound and stops the traverse.
Otherwise, the algorithm recursively traverses to child nodes of the BVHs.

\begin{algorithm}[h]
  \caption{$p_{col} = $PCD\_WS($T_A$, $T_B$, $T_{WS}$, $\delta$) \\: Compute the collision probability between polyhedra $A$ and $B$, given precomputed BVHs for polyhedra $T_A$ and $T_B$, a BVH $T_{WS}$ for the weighted samples, and a confidence level $\delta$.}
  \label{alg:pcd_ws}
  \begin{algorithmic}[1]
    \REQUIRE BVHs $T_A$, $T_B$, $T_{WS}$, confidence level $\delta$
    \ENSURE Upper bound on collision probability $p_{col}$
    \STATE $V_A$ = $T_A$.root
    \STATE $V_B$ = $T_B$.root
    \STATE $V_{WS}$ = $T_{WS}$.root
    \IF {IsNotCollision($V_{WS}$, $-V_A \bigoplus V_B$)}  \label{line:pcd_ws:overlap1}
      \RETURN $0$  \label{line:pcd_ws:overlap2}
    \ENDIF
    \IF {$T_{WS}$.sum\_weights $<$ $\delta$ or $V_A$, $V_B$, $V_{WS}$ are leaf nodes}  \label{line:pcd_ws:confidence1}
      \RETURN $p_{col}$ = $T_{WS}$.sum\_weights  \label{line:pcd_ws:confidence2}
    \ENDIF
    \IF {$V_A$ has children and $Vol(V_A) \geq Vol(V_B)$ and $Vol(V_A) \geq Vol(V_{WS})$}
      \RETURN $p_{col} = \sum\limits_{c_A : \text{child}}$PCD\_WS($c_A$, $T_B$, $T_{WS}$, $\delta$)
    \ELSIF{$V_B$ has children and $Vol(V_B) \geq Vol(V_A)$ and $Vol(V_B) \geq Vol(V_{WS})$}
      \RETURN $p_{col} = \sum\limits_{c_B : \text{child}}$PCD\_WS($T_A$, $c_B$, $T_{WS}$, $\delta$)
    \ELSE
      \RETURN $p_{col} = \sum\limits_{c_{WS} : \text{child}}$PCD\_WS($T_A$, $T_B$, $c_{WS}$, $\delta$)
    \ENDIF
  \end{algorithmic}
\end{algorithm}

Algorithm~\ref{alg:pcd_ws} summarizes the probabilistic collision detection for weighted samples as error distributions.
The algorithm is given with three precomputed bounding volume hierarchies.
Lines~\ref{line:pcd_ws:overlap1}-\ref{line:pcd_ws:overlap2} check for collisions between the bounding volumes and returns $0$ when there is no overlap.
Lines~\ref{line:pcd_ws:confidence1}-\ref{line:pcd_ws:confidence2} check whether the upper bound of collision is less than the confidence level $\delta$.
If the condition is satisfied, it returns the sum of weights instead of further traversing the tree.

\begin{algorithm}[h]
  \caption{$p_{col} = $PCD\_TGMM($T_A$, $T_B$, $T_{TGMM}$, $\delta$) \\: Compute the collision probability between polyhedra $A$ and $B$, given precomputed BVHs for polyhedra $T_A$ and $T_B$, a BVH $T_{TGMM}$ for the TGMMs, and a confidence level $\delta$.}
  \label{alg:pcd_tgmm}
  \begin{algorithmic}[1]
    \REQUIRE BVHs $T_A$, $T_B$, $T_{TGMM}$, confidence level $\delta$
    \ENSURE Upper bound on collision probability $p_{col}$
    \STATE $V_A$ = $T_A$.root
    \STATE $V_B$ = $T_B$.root
    \STATE $V_{TGMM}$ = $T_{TGMM}$.root
    \IF {IsNotCollision($V_{TGMM}$, $-V_A \bigoplus V_B$)}  \label{line:pcd_tgmm:overlap1}
      \RETURN $0$  \label{line:pcd_tgmm:overlap2}
    \ENDIF
    \IF {$T_{TGMM}$.sum\_weights $<$ $\delta$}  \label{line:pcd_tgmm:confidence1}
      \RETURN $p_{col}$ = $T_{TGMM}$.sum\_weights  \label{line:pcd_tgmm:confidence2}
    \ENDIF
    \IF {$T_{TGMM}$ is a leaf node}  \label{line:pcd_tgmm:detailed_computation1}
      \RETURN PCD\_TG($T_A$, $T_B$, $T_{TGMM}$.$\mu$, $T_{TGMM}$.$\Sigma$, $T_{TGMM}$.$r$, $\delta$)  \label{line:pcd_tgmm:detailed_computation2}
    \ENDIF
    \IF {$V_A$ has children and $Vol(V_A) \geq Vol(V_B)$ and $Vol(V_A) \geq Vol(V_{TGMM})$}
      \RETURN $p_{col} = \sum\limits_{c_A : \text{child}}$PCD\_TGMM($c_A$, $T_B$, $T_{TGMM}$, $\delta$)
    \ELSIF{$V_B$ has children and $Vol(V_B) \geq Vol(V_A)$ and $Vol(V_B) \geq Vol(V_{TGMM})$}
      \RETURN $p_{col} = \sum\limits_{c_B : \text{child}}$PCD\_TGMM($T_A$, $c_B$, $T_{TGMM}$, $\delta$)
    \ELSE
      \RETURN $p_{col} = \sum\limits_{c_{TGMM} : \text{child}}$PCD\_TGMM($T_A$, $T_B$, $c_{TGMM}$, $\delta$)
    \ENDIF
  \end{algorithmic}
\end{algorithm}

Algorithm~\ref{alg:pcd_tgmm} corresponds to the combined algorithm of Algorithm~\ref{alg:pcd_ws} and~\ref{alg:pcd_tg}, when the input error distribution is a TGMM.
The first collision check is performed on Line~\ref{line:pcd_tgmm:overlap1}, similar to the case for the  Weighted Samples, returning $0$ in Line~\ref{line:pcd_tgmm:overlap2} if there is no overlap.
Because the maximum possible value of collision probability for a single TG is its weight, the confidence level check in Line~\ref{line:pcd_tgmm:confidence1} is also performed similarly, comparing the sum of weights of the TGs with the confidence level $\delta$.
In Line~\ref{line:pcd_tgmm:detailed_computation1}, when the traversal reaches a leaf node of the BVH for TGMMs, it returns the collision probability obtained from \texttt{PCD\_TG} (Algorithm~\ref{alg:pcd_tg}).
Otherwise, the BVH traversal continues until it reaches the leaf nodes of BVHs.

\begin{figure}[ht]
  \centering
  \subfloat[][]
  {
    \includegraphics[width=0.4\linewidth]{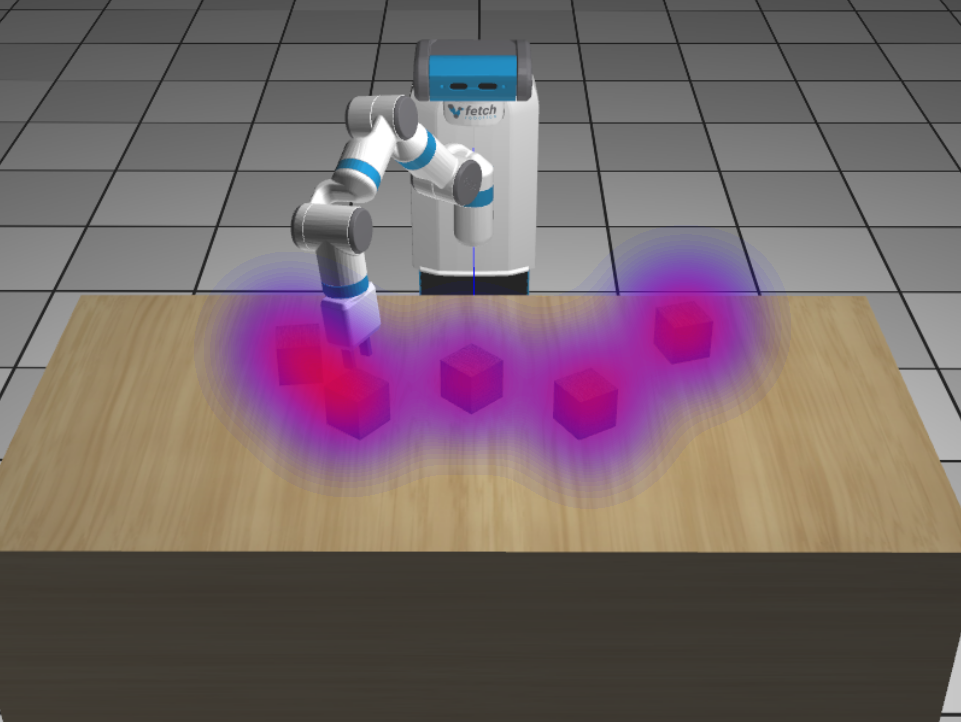}
  }
  \subfloat[][]
  {
    \includegraphics[width=0.4\linewidth]{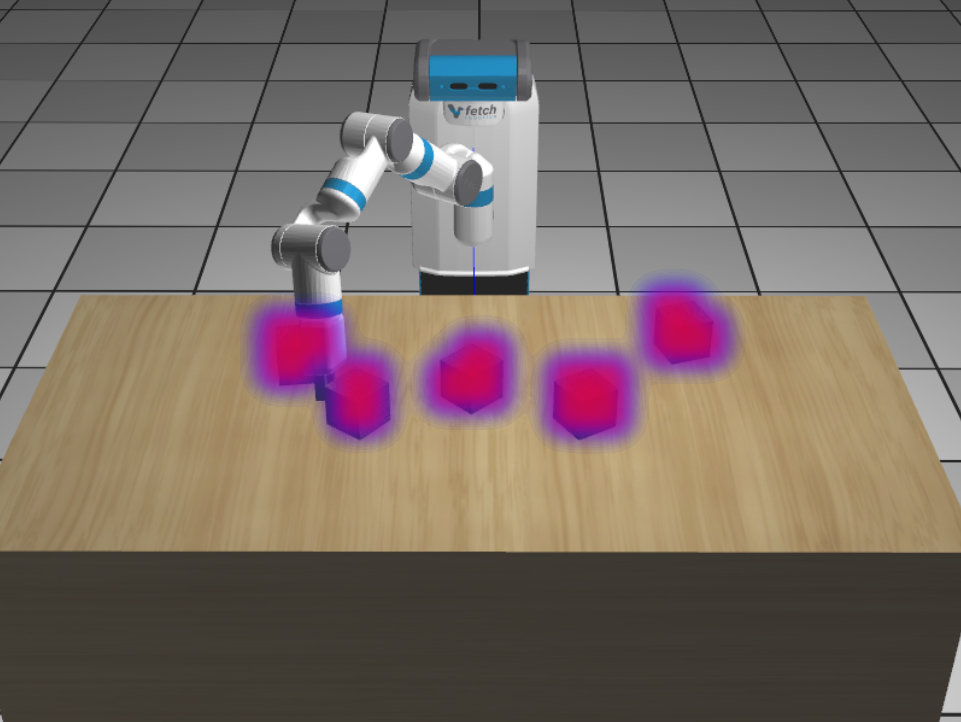}
  }
  \caption{The robot trajectory with probabilistic collision detection through narrow passages. (a) The wood block obstacles are captured by RGBD sensors and its positional errors are modeled with Gaussian distributions. In this case, the planner is unable to compute a path in the narrow passage because of the conservative error bounds on the collision probability. (b) Robot motion planning with probabilistic collision detection under error distribution in the form of TGMM can generate a collision-free robot trajectory that passes through the narrow passage, due to tight bounds.
  }
  \label{fig:pcd_robot}
  \vspace*{-0.15in}
\end{figure}

\subsection {Motion Planning}
Our probabilistic collision detection algorithm has been integrated with optimization-based motion planning and integrated with the 7-DOF Fetch arm (see Figure 1).  The tighter bounds on the collision probability enable us to find collision-free paths in tight spaces or narrow passages.

Tabel~\ref{table:motion_planning} shows the results of robot motion planning in the narrow passage scenario.
Exact collision detection between known shapes without uncertainties (CD-Obstacles) always gives a correct collision state, thus $0 \%p$ in collision probability over-estimation measurements. The robot motion planning without uncertainties operates perfectly, passing all narrow passages, generating non-collision trajectories, and keeping the closest distance between the end-effector and the table.
However, under sensor uncertainties, motion planning with the exact collision detection between the robot and the point clouds (CD-Points) works poorly. Due to the sensor uncertainties, a high error at a pixel may affect the collision detection query, reporting in-collision though the robot is not in collision, and vice versa. Due to this reason, the average and the standard deviation of collision probability over-estimation is the highest among other algorithms. This also impacts on the worst performance on the number of passages, success rate, and the distance measurements.
Motion planning algorithms with probabilistic collision detection algorithms show better performances than the exact collision detection algorithm under the environment with sensor uncertainties.
Compared to PCD-Gaussian, the probabilistic collision detection algorithms with non-Gaussian generally show better performances in the metrics in the table.
PCD-WS and PCD-TGMM run slower because of the complex shape of error distributions, but compute collision probability more accurately.
PCD-TGMM shows the best performance on collision probability estimation, the number of successful passages, successful avoiding obstacles, and the distance between the table surface and the end-effector.

Figure~\ref{fig:pcd_robot}  shows the results computed using our motion planner with different combination of  probabilistic collision detection algorithms on a benchmark that consists of wooden blocks. This wooden block has four narrow passages and have varying sizes.  If the collision probability computation results in a higher probability bound, the resulting planner concludes that there is no collision-free trajectory.
As a result, the planner is not able to compute a collision-free trajectory for all four cases, which is shown in terms of number of passes (i.e. \# Passages) and  the success rate in Table. 1.
We use this benchmark to evaluate the benefits of computing tighter probabilistic collision detection bounds.
We use an optimization-based planner, ITOMP~\cite{Park:2012:ICAPS}, which repeatedly refines the trajectory while interleaving the execution and motion planning for dynamic scenes. We handle three types of constraints: smoothness constraint, static obstacle collision-avoidance, and dynamic obstacle collision avoidance. 
In Figure~\ref{fig:pcd_robot} (a), the robot motion is planned with the probabilistic collision algorithm used for Gaussian error distribution.
However, this planner fails in the configurations corresponding to the narrow passage, even though there is sufficient space for the robot to pass through.
Because of the infinite domain of a Gaussian distributions, the non-zero collision probability prevents the robot to pass through, as it conveys a possible collision.
In Figure~\ref{fig:pcd_robot} (b), the probabilistic collision detection algorithm with TGMM is used in the motion planner.
Different from the Gaussian distribution case, the planner can compute a path through the narrow passages.
This is because the error distributions have truncation boundaries and the collision detection algorithm yields collision probability of under $5\%$. Overall, our probabilistic collision detection algorithm provides better bounds than prior methods.